\documentclass[11pt,a4paper]{article}
\usepackage[hyperref]{acl2021}
\usepackage{times}
\usepackage{latexsym}

\usepackage{graphicx}
\usepackage{CJKutf8}
\usepackage{multirow}
\usepackage{array}
\usepackage{tabularx}
\usepackage{makecell}
\usepackage{amsmath} 
\usepackage{amssymb}
\usepackage{tablefootnote}
\usepackage{subcaption}
\usepackage{diagbox}
\usepackage{algorithmic}
\usepackage{algorithm}
\usepackage{booktabs}
\usepackage{xcolor}
\usepackage{color, soul} %用color, 和 soul 包
\usepackage{colortbl}
\usepackage{makecell}
\usepackage{pgfplots}
\usepackage{arydshln}

\usepackage{microtype}

\aclfinalcopy % Uncomment this line for the final submission
\newcolumntype{L}[1]{>{\raggedright\arraybackslash}p{#1}}
\newcolumntype{C}[1]{>{\centering\arraybackslash}p{#1}}
\newcolumntype{R}[1]{>{\raggedleft\arraybackslash}p{#1}}

\DeclareSymbolFont{extraup}{U}{zavm}{m}{n}
\DeclareMathSymbol{\varheart}{\mathalpha}{extraup}{86}
\DeclareMathSymbol{\vardiamond}{\mathalpha}{extraup}{87}

\title{Grammatical Error Correction as GAN-like Sequence Labeling}
\author{Kevin Parnow$^{1,2,3,\dag}$, Zuchao Li$^{1,2,3,\dag}$, and Hai Zhao$^{1,2,3}$\thanks{$\ $  Corresponding author. $^\dag$ These authors made equal contribution. This work was supported by Huawei Noah's Ark Lab and funded by the National Key Research and Development Program
of China (No. 2017YFB0304100), the Key Projects of National Natural Science Foundation of China (U1836222 and
61733011), the Huawei-SJTU long term AI project, Cutting-edge
machine reading comprehension and language model.}\\
$^{1}$Department of Computer Science and Engineering, Shanghai Jiao Tong University \\
	$^{2}$Key Laboratory of Shanghai Education Commission for Intelligent Interaction \\ and Cognitive Engineering, Shanghai Jiao Tong University, Shanghai, China\\
	$^{3}$MoE Key Lab of Artificial Intelligence, AI Institute, Shanghai Jiao Tong University \\
  {\tt \small \{parnow, charlee\}@sjtu.edu.cn, zhaohai@cs.sjtu.edu.cn}}

\date{}

\begin{document}
\maketitle

\begin{abstract}

In Grammatical Error Correction (GEC), sequence labeling models enjoy fast inference compared to sequence-to-sequence models; however, inference in sequence labeling GEC models is an iterative process, as sentences are passed to the model for multiple rounds of correction, which exposes the model to sentences with progressively fewer errors at each round. 
Traditional GEC models learn from sentences with fixed error rates. 
Coupling this with the iterative correction process causes a mismatch between training and inference that affects final performance. 
In order to address this mismatch, we propose a GAN-like sequence labeling model, which consists of a grammatical error detector as a discriminator and a grammatical error labeler with Gumbel-Softmax sampling as a generator. 
By sampling from real error distributions, our errors are more genuine compared to traditional synthesized GEC errors, thus alleviating the aforementioned mismatch and allowing for better training. 
Our results on several evaluation benchmarks demonstrate that our proposed approach is effective and improves the previous state-of-the-art baseline.

\end{abstract}

\section{Introduction}

Sequence-to-sequence neural solutions \cite{DBLP:conf/ialp/ParnowLZ20} have been quite successful in comparison to their statistical counterparts \cite{DBLP:conf/nips/SutskeverVL14}, but these approaches suffer from a couple key problems, which has given rise to sequence labeling approaches for GEC \cite{omelianchuk-etal-2020-gector}. Such approaches task models with generating a list of labels to classify the grammatical errors in a sentence before correcting these errors.

Sequence labeling approaches have recently gained popularity in GEC and are currently state-of-the-art. 
One typical aspect of sequence labeling approaches \cite{he-etal-2018-syntax,li-etal-2018-unified} is labeling and correcting sentences through an iterative process. As successive edits will depend on how other errors are corrected in a sentence, using an iterative process and correcting only the most salient errors in each round allows models to achieve better performance; however, because of this process, models are tasked with handling sentences with varying rates of errors, as during each round of inference for a given sentence, a model encounters a sentence with progressively fewer errors. This of course causes an exposure bias problem, as the training data does not match the test data, and suggests that providing the model with training data with varying error rates will lead to better performance. 

To combat this exposure bias, we propose a new approach for training a sequence labeling GEC model that draws from GANs \cite{DBLP:conf/nips/GoodfellowPMXWOCB14}, which consist of a generator that generates increasingly realistic fake inputs and a discriminator that is tasked with differentiating these fake inputs from real inputs. 
Other GEC works like \cite{raheja-alikaniotis-2020-adversarial} directly used GANs to produce grammatically correct sentences given grammatically incorrect ones. This contrasts our work, which uses aspects of a GAN to enhance the training process rather than using a GAN itself as the correcting model. 
Our model consists of three components: an encoder, a Grammatical Error Detector, and a Grammatical Error Labeler. By sampling from the error distribution in the error labeler, our model can synthesize sentences with new errors creating new sentence pairs for further training data. As a result, our Detector continually improves its ability to detect errors and essentially acts as a discriminator of errors, and our Labeler continually improves the authenticity of its error distribution and becomes a better generator of errors.
This process allows us to counter the exposure bias problem sequence labeling GEC models face because in addition to allowing us to generate new errorful sentences whose errors are increasingly representative of those in real data, we can also use control parameters to set the error rates of these sentences and accommodate our iterative inference process.

\section{Our Approach}

\begin{figure}[t]
	\centering
	\includegraphics[width=.49\textwidth]{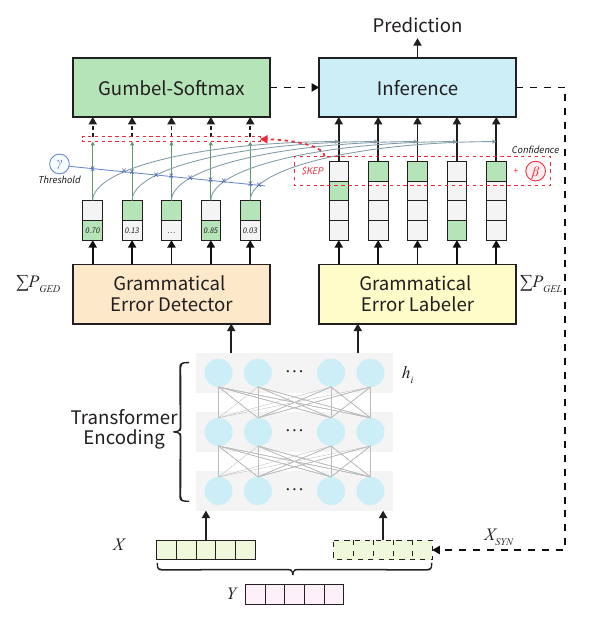}
	\caption{An overview of our model.}
	\vspace{-20pt}
	\label{fig:overview}
\end{figure}

We formulate the GEC task as a problem of sequence labeling and create a neural sequence labeling model based on a deep pre-trained Transformer encoder to deal with this problem. Inspired by the work of \cite{omelianchuk-etal-2020-gector}, our full model's overall architecture diagram is shown in Figure \ref{fig:overview}. 
There are three main components in our basic neural GEC model: a deep pre-trained Transformer Encoder, a Grammatical Error Detector, and a Grammatical Error Labeler.
To accommodate our new GAN-like training process, we add a Gumbel-softmax sampling component to the basic GEC model.

\subsection{Background and Notation}
First, in training, given incorrect input sentence $X = x_1, x_2, ..., x_n$ and its corrected version $X_c = y_1, y_2, ..., y_m$, the model predicts a corrective label sequence $T = t_1, t_2, ..., t_n$ by minimizing the token-level Levenshtein distance on the span-based alignments of $X$ and $X_c$.
The corrective label set is given as $\mathcal{T} = \{\textnormal{\$KEP}, \textnormal{\$DEL}, \textnormal{\$APP}, \textnormal{\$REP}\} \cup \{\textnormal{\$CAS}, \textnormal{\$MRG}, \textnormal{\$SPL}, \textnormal{\$NNUM}, \textnormal{\$VFORM}\}$, in which the first set consists of the basic text editing transformation operations and the second consists of \textit{g}-transformations as defined by \cite{omelianchuk-etal-2020-gector} for GEC\footnote{The label set here only presents the transformations' basic names. Some transformations require additional parameters because they are context-specific and thus have many different versions.}.
Aligning sentences using these transformations in pre-processing, reduces what would be a sequence generation task that handles unequal source-target lengths to a set of label classification problems. In this formulation, the neural sequence labeling model trains to optimize the input sequence's negative log-likelihood loss for an input sequence:
$$\mathcal{J}(\theta) = - \sum_{i=1}^{n} \log p(t_i | x, \theta),$$
where $p$ is the conditional probability that the model outputs at each position $i$.

\subsection{Deep Pre-trained Transformer Encoder}
As in most neural sequence labeling models \cite{ma-hovy-2016-end}, a neural encoder such as a BiLSTM \cite{hochreiter1997long} or a Transformer \cite{vaswani2017attention,DBLP:journals/corr/abs-2102-05951} is used to extract context-aware features from the input sequence. 
Deep pre-trained language models such as BERT \cite{devlin-etal-2019-bert,DBLP:conf/aaai/0001WZLZZZ20}, RoBERTa \cite{liu2019roberta}, and XLNet \cite{DBLP:conf/nips/YangDYCSL19} have recently demonstrated the efficacy of Transformer models trained on large-scale unlabeled data in various NLP tasks.
We leveraged these very beneficial models by using a pre-trained language model as our encoder.
We define the contextualized features captured by the neural encoder as:
$$h_i = [\mathbf{Enc}(X)]_i,$$
where $\mathbf{Enc}$ represents the encoder, and $[\cdot]_i$ represents the output of the $i$-th position after encoding.

\subsection{Grammatical Error Detector and Labeler}
Next, we adopt a a Grammatical Error Detector (GED) to detect the presence of errors and a Grammatical Error Labeler (GEL) to predict detailed error labels. 
With these labels, corrections are applied to sentences, and this process is typically iterative, as some corrections may depend on others, and applying corrections only once may not be enough to fully correct the sentence. 
During iterative correction, the model needs to assess at each round whether more correction is required. 
To this end, we use the GED to determine the degree of error for an entire sentence and control the iterative correction process.

Specifically, we use a binarization $Y_b$ of the corrective labels $Y$ as the training target of the GED and use $Y$ as the training target of the GEL. 
To obtain label probabilities grammatical error detection and labeling, two linear layers with softmax layers are appended to the encoder:
$$P_{\textnormal{GED}}^i = \mathbf{softmax}(\mathbf{MLP}_{\textnormal{GED}}(h_i)),$$
$$P_{\textnormal{GEL}}^i = \mathbf{softmax}(\mathbf{MLP}_{\textnormal{GEL}}(h_i)).$$

The binary classification probabilities in the GED output do not necessarily control the inference process's iterations. 
Rather, after using the GEL error label probabilities as thresholds for sentence positions, we also use the sum of these probabilities as a threshold for attempting another round of correction on the whole sentence.
The model continues correcting the sentence until either it reaches a preset maximum number of iterations or no longer satisfies the following condition: $\sum_{i} [P_{\textnormal{GED}}^i]_{err=1} > \gamma$, where $\gamma$ is the minimum error probability threshold for a sentence.

Additionally, since GEC usually corrects a small portion of a sentence (and there are therefore no errors in most of the input), the corrective label prediction task is an imbalanced classification problem.
We alleviate this imbalance classification issue by taking advantage of this prior knowledge and adding a fixed and preset confidence $\beta$ to the label $\textnormal{\$KEP}$ to keep a position unchanged when applying corrections: 
$$[P_{\textnormal{GEL}}^i]_{\textnormal{\$KEP}} = [P_{\textnormal{GEL}}^i]_{\textnormal{\$KEP}} + \beta.$$

\subsection{GAN-like Sequence Labeling Training}

While we adopt sequence labeling instead of sequence-to-sequence modeling in this paper and therefore avoid the exposure bias problem caused by left-to-right sequence generation, our model still faces exposure bias because of the iterative correction process, which, through its iterative correction process, tasks the model with handling much more varied error rates in inference compared to in training, where it handles static data and does not use multiple-round corrections. 
To address this issue, we borrow the idea of a GAN \cite{DBLP:conf/nips/GoodfellowPMXWOCB14} and propose a GAN-like iterative training approach for a sequence labeling GEC model. GANs, whose training objective can be formulated as a minimax game between a generator that creates increasingly realistic fake outputs and a discriminator that must differentiate these outputs from their real counterparts, have been suggested for sequence-to-sequence text generation \cite{DBLP:conf/iclr/LiWCUS0Z20,DBLP:conf/iclr/0001C0USLZ20,li-etal-2018-seq2seq} as they do not suffer from exposure bias.

\begin{algorithm}[]%[t]%[!h]
	\small
	\centering
	\caption{\small GAN-like Sequence Labeling Training}\label{alg:gan}
	\begin{algorithmic}[1]
		\REQUIRE Genuine GEC parallel dataset $\mathcal{D} = \{(\textbf{X}, \textbf{Y})\}$ \\
		Synthesized GEC parallel dataset $\mathcal{D}_{\textnormal{SYN}} = \{\}$ \\
		Number of training stages $\mathcal{N}$ \\
		Number of training epochs $M$\\
		Sentence error probability threshold $\gamma$ \\
		Additional confidence $\beta$ for label {\tt \$KEP} \\
		\FOR{$i$ in 1, ..., $N$}
		    \STATE Initialize model parameters from previous training stage $\theta_{\textnormal{i}} \leftarrow \theta_{\textnormal{i-1}}$ when $i > 1$
		    \FOR{$j$ in 1, ..., $M$}
    	        \FOR{$k$ in 1, ..., $|\mathcal{D} \cup \mathcal{D}_{\textnormal{SYN}}|$}
        		\STATE Encode each sentence $\textbf{X}_k$ as $\textbf{H}_k$ %using the encoder 
    	        \STATE $P_{\textnormal{GED}}^k = \mathbf{Softmax}(\textnormal{MLP}_{\textnormal{GED}}(\textbf{H}_k))$
    	        \STATE $P_{\textnormal{GEL}}^k = \mathbf{Softmax}(\textnormal{MLP}_{\textnormal{GEL}}(\textbf{H}_k))$
    	        \STATE $loss_{\textnormal{GED}} = \mathbf{CrossEntropy}(P_{\textnormal{GED}}^k, \textbf{Y}_{err}^k)$
    	        \STATE $loss_{\textnormal{GEL}} = \mathbf{CrossEntropy}(P_{\textnormal{GEL}}^k, \textbf{Y}_{label}^k)$
    	        \STATE $loss = loss_{\textnormal{GED}} + loss_{\textnormal{GEL}}$
    	        \STATE Update the model parameter $\theta_i$ with $loss$
    	        \ENDFOR
		    \ENDFOR
		    \STATE $\mathcal{D}_{\textnormal{SYN}} = \{\}$
		    \FOR{$k$ in 1, ..., $|\mathcal{D}|$}
        		\STATE Encode each sentence $\textbf{X}_k$ as $\textbf{H}_k$ %using the Transformer-based encoder 
                \STATE $P_{\textnormal{GED}}^k = \mathbf{Softmax}(\textnormal{MLP}_{\textnormal{GED}}(\textbf{H}_k))$
                \STATE $P_{\textnormal{GED}}^k = \sum [P_{\textnormal{GED}}^k]_{err=1} > \gamma$
                \STATE $P_{\textnormal{GEL}}^k = \mathbf{Softmax}(\textnormal{MLP}_{\textnormal{GEL}}(\textbf{H}_k))$
                \STATE $[P_{\textnormal{GEL}}^k]_{\tt \$KEP} = [P_{\textnormal{GEL}}^k]_{\tt \$KEP} + \beta$
                \STATE $P_{\textnormal{GEL}}^k = \mathbf{GumbelSoftmax}(P_{\textnormal{GEL}}^k)$
                \STATE Use $P_{\textnormal{GED}}^k$ and $P_{\textnormal{GEL}}^k$ to produce sampled sequence $\mathbf{X}_{\textnormal{SYN}}^k$
                \STATE $\mathcal{D}_{\textnormal{SYN}} = \mathcal{D}_{\textnormal{SYN}} \cup \{(\mathbf{X}_{\textnormal{SYN}}^k, \mathbf{Y}_k)\}$
            \ENDFOR
	\ENDFOR
	\end{algorithmic}
\end{algorithm}

\begin{table*}[t]
    \centering
    \small
    \begin{tabular}{lccccccccccc}
    \toprule
    \multirow{2}{*}{\textbf{GEC system}} & \multirow{2}{*}{\textbf{Ens.}} & & \multicolumn{3}{c}{\textbf{CoNLL-2014 (test)}} & & \multicolumn{3}{c}{\textbf{BEA-2019 (test)}} && \multicolumn{1}{c}{\textbf{JFLEG (test)}}\\ 
    \cmidrule{4-6} \cmidrule{8-10} \cmidrule{12-12}& & & \textbf{P} &  \textbf{R}  & $\mathbf{F_{0.5}}$ & & \textbf{P} &  \textbf{R}  & $\mathbf{F_{0.5}}$ & & $\mathbf{GLEU}$ \\ 
    \midrule    
    \cite{zhao2019improving}& & &   67.7   &  40.6   &   59.8   & &   $-$   &  $-$   &   $-$  && $-$\\ 
    \cite{awasthi-etal-2019-parallel}&&&   66.1   &  43.0    &   59.7   &&   $-$   &  $-$    &   $-$ && 60.3 \\ 
    \cite{kiyono2019empirical}&&&   67.9  &  44.1   &  61.3  & & 65.5 &  59.4   &   64.2  && 59.7 \\ 
    \cite{kaneko-etal-2020-encoder}&&& 69.2 & 45.6 & 62.6 && 67.1 & 60.1 & 65.6 && 61.3\\
    \midrule
    \cite{lichtarge-etal-2019-corpora}&\checkmark&& 66.7 & 43.9 & 60.4 && $-$ & $-$ & $-$ && \bf 63.3 \\
    \cite{zhao2019improving}&\checkmark&&   71.6 &  38.7   &  61.2   &&   $-$   &  $-$   &   $-$  && 61.0 \\ 
    \cite{awasthi-etal-2019-parallel}&\checkmark&&   68.3   &  43.2    &   61.2  & &   $-$   &  $-$    &   $-$ && 61.0 \\ 
    \cite{kiyono2019empirical}&\checkmark&&   72.4  &  46.1   &  65.0  && 74.7 &  56.7   &   70.2   && 61.4\\ 
    \cite{kantor2019learning}&\checkmark&&   $-$ &  $-$   & $-$  && 78.3 &  58.0   &   \bf 73.2   && $-$ \\ 
    \cite{kaneko-etal-2020-encoder} &\checkmark && 72.6 & \bf 46.4 & 65.2 && 72.3 & \bf 61.4 & 69.8 && 62.0 \\
    \midrule
    Baseline (BERT-base) &&& 72.1 & 42.0 & 63.0 && 71.5  & 55.7 &  67.6 & & 60.1 \\ 
    \quad \bf+GST & &  & 72.6 & 42.5 & 63.6 & & 71.9 & 55.9 & 68.0 & & 60.5 \\ \hdashline
    Baseline (RoBERTa-base) &&& 73.9 &  41.5 & 64.0 && 77.2 & 55.1 & 71.5  && 60.6 \\ 
    \quad\bf+GST & & & 74.1 & 42.2 & 64.4 & & 77.5 & 55.7 & 71.9 &  & 60.9 \\ \hdashline
    Baseline (XLNet-base) &&& 77.5 &  40.1   & 65.3 && 79.2    & 53.9 & 72.4 && 61.5 \\ 
    \quad\bf+GST & & & \bf 78.4 & 39.9 & \bf 65.7 & & \bf 79.4 & 54.5 & 72.8 & & 61.8 \\
    \bottomrule
    \end{tabular}
    \caption{\label{results-table} Comparison of GEC models. The baseline comes from the model released by \cite{omelianchuk-etal-2020-gector}.}
\end{table*}

In our model, the GEL module can be considered a discriminator, as it must differentiate whether tokens are erroneous, and by adding a sampling module to the GED module, we can create a generator that outputs grammatical errors (rather than corrections) that are increasingly realistic. We can then pair these sampling outputs with their golden sequence in the training dataset to create new training samples. This trains the model with more samples and more varied errors and alleviates the exposure bias issue. Separate cross-entropy losses are calculated for the Grammatical Error Detector and Labeler, and we detail the whole algorithm for our training process in Algorithm \ref{fig:alg}.

\section{Detailed Training Process} \label{fig:alg}

To synthesize new errors based on a genuine grammatical error distribution, we add a sampling module to a trained GED module. Specifically, we use Gumbel-softmax sampling, a simple and efficient way to draw samples $z$ from a categorical distribution with class probabilities $P_{\textnormal{GEL}}$ using the Gumbel-Max trick \cite{gumbel1954statistical,DBLP:conf/nips/MaddisonTM14}:
\begin{equation} \label{eq:gumbel}
z = \verb|one_hot|\left(\mathbf{argmax}_{j}{\left[ g_j + \log [P_{\textnormal{GEL}}^i]_j \right]}\right)
\end{equation}
where $g_1...g_j$ are i.i.d samples drawn from $\text{Gumbel}(0,1)$\footnote{The $\text{Gumbel}(0,1)$ distribution can be sampled using inverse transform sampling by drawing $u \sim \text{Uniform}(0, 1)$ and computing $g= -\log(-\log(\text{u}))$. }.
We use the softmax function as a continuous, differentiable approximation to $\mathbf{argmax}$:
\begin{equation}
[y_i]_{k} = \frac{\text{exp}((\log([P_{\textnormal{GEL}}^i]_{k})+g_k)/\tau)}{\sum_{j=1}^{|C|} \text{exp}((\log([P_{\textnormal{GEL}}^i]_j)+g_j)/\tau)},
\end{equation}
where $|C|$ is the number of classes, $\tau$ is the softmax temperature. Altering $\gamma$ and $\beta$ allows us to synthesize input samples of different error rates.

\section{Experiments}
\subsection{Setup}
To isolate our GAN-like Sequence Labeling Training (GST) approach, we use the same model setting and training details as in \cite{omelianchuk-etal-2020-gector}. The training data includes PIE's synthetic data \cite{awasthi-etal-2019-parallel}, NUCLE \cite{dahlmeier-etal-2013-building}, Lang-8 \cite{tajiri-etal-2012-tense}, FCE \cite{yannakoudakis-etal-2011-new}, Cambridge Learner Corpus (the publicly available portion) \cite{nicholls2003cambridge}, and WI+LOCNESS \cite{bryant-etal-2019-bea}. Our models are evaluated on the test sets of CoNLL-2014 \cite{ng-etal-2014-conll}, BEA-2019 \cite{bryant-etal-2019-bea}, and JFLEG \cite{napoles-etal-2017-jfleg} with the official $M^2$ \cite{dahlmeier-ng-2012-better}, ERRANT \cite{bryant-etal-2017-automatic}, and GLEU\cite{napoles-etal-2015-ground} scorers, respectively.

\begin{table}[]
    \centering
    \small
    \begin{tabular}{c|ccc}
         \toprule
         \multirow{2}{*}{\bf Sampling} &   \multicolumn{3}{c}{\textbf{CoNLL-2014 (test)}} \\
         & \textbf{P} &  \textbf{R}  & $\mathbf{F_{0.5}}$ \\
         \midrule
         \textit{Random} & 74.3 & 40.2 & 63.5 \\
         \textit{GumbelSoftmax} & 78.4 & 39.9 & 65.7 \\
         \textit{Multinomial} & 78.1 & 39.9 & 65.5 \\
         \bottomrule
    \end{tabular}
    \caption{Comparing the effects of different sampling distributions.}
    \label{tab:ablation_sampling}
\end{table}

\subsection{Results and Analysis}

Our results on the three test datasets are listed in Table \ref{results-table}. 
Our baseline model achieves the best single model CoNLL-2014 $F_{0.5}$, BEA-2019 $F_{0.5}$, and JFLEG GLEU scores, showing that the baseline we use is very strong.
The results on the three benchmarks are further improved using the GST approach, which demonstrates that the GST approach can effectively alleviate the exposure bias issue. 
With GST, we achieved new best results on the CoNLL-2014 test dataset, surpassing ensemble methods while only using a single model. 

In order to illustrate the benefits of sampling using Gumbel-Softmax, we replaced it with random sampling and Multinomial. The comparison is shown in Table \ref{tab:ablation_sampling}.
Random sampling actually hampers performance, which shows that synthetic sentences not based on a genuine error distribution do not alleviate exposure bias. Both GumbelSoftmax and Multinomial, which use a genuine error distribution, improve the model, though GumbelSoftmax appears to be more suitable for sampling in sequence labeling modeling.

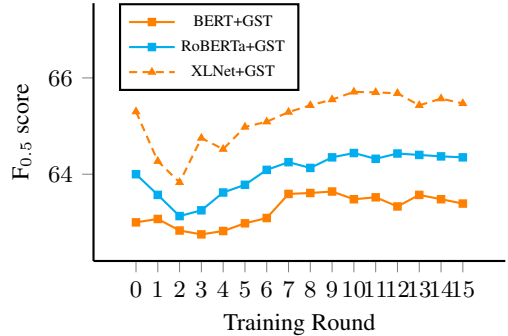
\begin{figure}[]
	\centering
		\setlength{\abovecaptionskip}{0pt}
		\begin{center}
			\pgfplotsset{height=5cm,width=8.5cm,compat=1.14,every axis/.append style={thick},every axis legend/.append style={ at={(0.5,1)}},legend columns=1}
			\begin{tikzpicture}
			\tikzset{every node}=[font=\small]
			\begin{axis}
			[width=7cm,enlargelimits=0.13, tick align=outside, xticklabels={ $0$, $1$, $2$, $3$, $4$, $5$, $6$, $7$, $8$, $9$, $10$, $11$, $12$, $13$, $14$, $15$},
            axis y line*=left,
            xtick={0,1,2,3,4,5,6,7,8,9,10,11,12,13,14,15},
            ylabel={F$_{0.5}$ score},
            axis x line*=left, 
            ylabel style={align=left},xlabel={Training Round},font=\small,ymax=67]
			\addplot+ [sharp plot,mark=square*,mark size=1.2pt,mark options={solid,mark color=orange}, color=orange] coordinates
			{ (0,63.0)(1,63.07)(2,62.83)(3,62.75)(4,62.82)(5,62.98)(6,63.09)(7,63.59)(8,63.61)(9,63.64)(10,63.48)(11,63.52)(12,63.33)(13,63.57)(14,63.48)(15,63.39) };\label{plot_bb}
			\addlegendentry{\tiny BERT+GST}
			\addplot+ [sharp plot,mark=square*,mark size=1.2pt,mark options={solid,mark color=cyan}, color=cyan] coordinates
			{ (0,64.0)(1,63.57)(2,63.13)(3,63.25)(4,63.62)(5,63.78)(6,64.09)(7,64.25)(8,64.13)(9,64.35)(10,64.44)(11,64.32)(12,64.43)(13,64.40)(14,64.37)(15,64.35) };\label{plot_aa}
			\addlegendentry{\tiny RoBERTa+GST}
            \addplot+ [sharp plot,densely dashed,mark=triangle*,mark size=1.2pt,mark options={solid,mark color=orange}, color=orange] coordinates
			{ (0,65.3)(1,64.27)(2,63.83)(3,64.75)(4,64.52)(5,64.98)(6,65.09)(7,65.29)(8,65.43)(9,65.55)(10,65.71)(11,65.70)(12,65.68)(13,65.43)(14,65.57)(15,65.47) };
			\addlegendentry{\tiny XLNet+GST}
            \end{axis}
			\end{tikzpicture}
		\end{center}
	\caption{The GEC performance versus the the GST rounds on the CoNLL-2014 test set.}\label{fig:rounds} 
\end{figure}

In Figure \ref{fig:rounds}, we show how the performance changes with increasing rounds of GST training. 
In the first few rounds, due to the model's re-adaptation to new errors, there was a drop in performance on the test datasets; however, as the number of training rounds increased, performance on the test set gradually improved and finally stabilized.

\begin{table}[]
    \centering
    \small
    \begin{tabular}{c|ccc}
         \toprule
         \multirow{2}{*}{\bf Model} &   \multicolumn{3}{c}{\textbf{CoNLL-2014 (test)}} \\
         & \textbf{P} &  \textbf{R}  & $\mathbf{F_{0.5}}$ \\
         \midrule
         \textit{Baseline} & 72.1 & 42.0 & 63.0 \\
         \textit{Baseline + Extended Training} & 73.5 & 40.1 & 63.0 \\
         \textit{Intermediate Outputs} & 73.2 & 39.8 & 62.7 \\
         \textit{GST} & 72.6 & 42.5 & 63.6 \\
         \bottomrule
    \end{tabular}
    \caption{Comparing GST training with additional baselines.}
    \label{tab:inter}
\end{table}

\paragraph{Intermediate Outputs and Longer Training}
In this experiment, we explored using intermediate outputs from our iterative inference process as additional training outputs to highlight the impact of generating new erroneous sentence by sampling from the real error distribution with our GST approach. For this experiment, we use our baseline architecture. As seen in the results in Table \ref{tab:inter}, whereas GST leads to a 0.6 $F_{0.5}$ gain over the baseline, using intermediate training outputs paired with golden sentences for additional training actually leads to worse performance, yielding a 0.3 $F_{0.5}$ loss in comparison to the baseline. 

To confirm that GST's performance gain is not due to the added training time, we also train the baseline for a commensurate amount of additional steps but find that this does not have any effect on model performance. This experiment demonstrates that our model does bring improvement to the baseline without relying on additional training steps. We also note that as our model is not significantly different in size from our baseline, our improvement is also not brought about by simply using a larger model. 

\begin{table}[]
    \centering
    \small
    \begin{tabular}{c|ccc}
         \toprule
         \multirow{2}{*}{\bf Model} &   \multicolumn{3}{c}{\textbf{CoNLL-2014 (test)}} \\
         & \textbf{P} &  \textbf{R}  & $\mathbf{F_{0.5}}$ \\
         \midrule
         \textit{Baseline} & 72.1 & 42.0 & 63.0 \\
         \textit{GST} & 72.6 & 42.5 & 63.6 \\
         \hdashline
         \textit{Baseline w/o BERT} & 65.0 & 32.6 & 54.2 \\
         \textit{GST w/o BERT} & 64.9 & 35.6 & 55.7 \\
         \bottomrule
    \end{tabular}
    \caption{Evaluating GST without pre-trained language models.}
    \label{tab:prlm}
\end{table}

\paragraph{Performance with out Pre-trained Language Models}
We additionally explored the performance of our system in the absence of contextualized pre-trained language models. As we expected, these models make our model much more resilient to the exposure bias problem, and as seen in Table \ref{tab:prlm}, the improvement brought about GST is therefore much more evident. In comparison to the baseline, using GST brings an improvement of 1.5 $F_{0.5}$ points. 

\section{Conclusion}

In this paper, we studied the exposure bias problem GEC sequence labeling models face.
To alleviate this issue, we proposed a novel GAN-like training method for the GEC sequence labeling model. 
Through evaluation on three GEC benchmarks, we demonstrate that our novel training approach further improves a strong baseline model, illustrating the effectiveness of our training approach. 
Notably, with the help of pre-trained language models and our training approach, we achieved state-of-the-art results on the CoNLL-2014 benchmark.

\bibliographystyle{acl_natbib}
\bibliography{acl2021}

\begin{thebibliography}{36}
\expandafter\ifx\csname natexlab\endcsname\relax\def\natexlab#1{#1}\fi

\bibitem[{Awasthi et~al.(2019)Awasthi, Sarawagi, Goyal, Ghosh, and
  Piratla}]{awasthi-etal-2019-parallel}
Abhijeet Awasthi, Sunita Sarawagi, Rasna Goyal, Sabyasachi Ghosh, and Vihari
  Piratla. 2019.
\newblock \href {https://doi.org/10.18653/v1/D19-1435} {Parallel iterative edit
  models for local sequence transduction}.
\newblock In \emph{Proceedings of the 2019 Conference on Empirical Methods in
  Natural Language Processing (EMNLP-IJCNLP)}, pages 4260--4270, Hong Kong,
  China. Association for Computational Linguistics.

\bibitem[{Bryant et~al.(2019)Bryant, Felice, Andersen, and
  Briscoe}]{bryant-etal-2019-bea}
Christopher Bryant, Mariano Felice, {\O}istein~E. Andersen, and Ted Briscoe.
  2019.
\newblock \href {https://doi.org/10.18653/v1/W19-4406} {The {BEA}-2019 shared
  task on grammatical error correction}.
\newblock In \emph{Proceedings of the Fourteenth Workshop on Innovative Use of
  NLP for Building Educational Applications}, pages 52--75, Florence, Italy.
  Association for Computational Linguistics.

\bibitem[{Bryant et~al.(2017)Bryant, Felice, and
  Briscoe}]{bryant-etal-2017-automatic}
Christopher Bryant, Mariano Felice, and Ted Briscoe. 2017.
\newblock \href {https://doi.org/10.18653/v1/P17-1074} {Automatic annotation
  and evaluation of error types for grammatical error correction}.
\newblock In \emph{Proceedings of the 55th Annual Meeting of the Association
  for Computational Linguistics (Volume 1: Long Papers)}, pages 793--805,
  Vancouver, Canada. Association for Computational Linguistics.

\bibitem[{Dahlmeier and Ng(2012)}]{dahlmeier-ng-2012-better}
Daniel Dahlmeier and Hwee~Tou Ng. 2012.
\newblock \href {https://www.aclweb.org/anthology/N12-1067} {Better evaluation
  for grammatical error correction}.
\newblock In \emph{Proceedings of the 2012 Conference of the North {A}merican
  Chapter of the Association for Computational Linguistics: Human Language
  Technologies}, pages 568--572, Montr{\'e}al, Canada. Association for
  Computational Linguistics.

\bibitem[{Dahlmeier et~al.(2013)Dahlmeier, Ng, and
  Wu}]{dahlmeier-etal-2013-building}
Daniel Dahlmeier, Hwee~Tou Ng, and Siew~Mei Wu. 2013.
\newblock \href {https://www.aclweb.org/anthology/W13-1703} {Building a large
  annotated corpus of learner {E}nglish: The {NUS} corpus of learner
  {E}nglish}.
\newblock In \emph{Proceedings of the Eighth Workshop on Innovative Use of
  {NLP} for Building Educational Applications}, pages 22--31, Atlanta, Georgia.
  Association for Computational Linguistics.

\bibitem[{Devlin et~al.(2019)Devlin, Chang, Lee, and
  Toutanova}]{devlin-etal-2019-bert}
Jacob Devlin, Ming-Wei Chang, Kenton Lee, and Kristina Toutanova. 2019.
\newblock \href {https://doi.org/10.18653/v1/N19-1423} {{BERT}: Pre-training of
  deep bidirectional transformers for language understanding}.
\newblock In \emph{Proceedings of the 2019 Conference of the North {A}merican
  Chapter of the Association for Computational Linguistics: Human Language
  Technologies, Volume 1 (Long and Short Papers)}, pages 4171--4186,
  Minneapolis, Minnesota. Association for Computational Linguistics.

\bibitem[{Goodfellow et~al.(2014)Goodfellow, Pouget{-}Abadie, Mirza, Xu,
  Warde{-}Farley, Ozair, Courville, and
  Bengio}]{DBLP:conf/nips/GoodfellowPMXWOCB14}
Ian~J. Goodfellow, Jean Pouget{-}Abadie, Mehdi Mirza, Bing Xu, David
  Warde{-}Farley, Sherjil Ozair, Aaron~C. Courville, and Yoshua Bengio. 2014.
\newblock \href
  {https://proceedings.neurips.cc/paper/2014/hash/5ca3e9b122f61f8f06494c97b1afccf3-Abstract.html}
  {Generative adversarial nets}.
\newblock In \emph{Advances in Neural Information Processing Systems 27: Annual
  Conference on Neural Information Processing Systems 2014, December 8-13 2014,
  Montreal, Quebec, Canada}, pages 2672--2680.

\bibitem[{Gumbel(1954)}]{gumbel1954statistical}
Emil~Julius Gumbel. 1954.
\newblock \emph{Statistical theory of extreme values and some practical
  applications: a series of lectures}, volume~33.
\newblock US Government Printing Office.

\bibitem[{He et~al.(2018)He, Li, Zhao, and Bai}]{he-etal-2018-syntax}
Shexia He, Zuchao Li, Hai Zhao, and Hongxiao Bai. 2018.
\newblock \href {https://doi.org/10.18653/v1/P18-1192} {Syntax for semantic
  role labeling, to be, or not to be}.
\newblock In \emph{Proceedings of the 56th Annual Meeting of the Association
  for Computational Linguistics (Volume 1: Long Papers)}, pages 2061--2071,
  Melbourne, Australia. Association for Computational Linguistics.

\bibitem[{Hochreiter and Schmidhuber(1997)}]{hochreiter1997long}
Sepp Hochreiter and J{\"u}rgen Schmidhuber. 1997.
\newblock Long short-term memory.
\newblock \emph{Neural computation}, 9(8):1735--1780.

\bibitem[{Kaneko et~al.(2020)Kaneko, Mita, Kiyono, Suzuki, and
  Inui}]{kaneko-etal-2020-encoder}
Masahiro Kaneko, Masato Mita, Shun Kiyono, Jun Suzuki, and Kentaro Inui. 2020.
\newblock \href {https://doi.org/10.18653/v1/2020.acl-main.391}
  {Encoder-decoder models can benefit from pre-trained masked language models
  in grammatical error correction}.
\newblock In \emph{Proceedings of the 58th Annual Meeting of the Association
  for Computational Linguistics}, pages 4248--4254, Online. Association for
  Computational Linguistics.

\bibitem[{Kantor et~al.(2019)Kantor, Katz, Choshen, Cohen-Karlik, Liberman,
  Toledo, Menczel, and Slonim}]{kantor2019learning}
Yoav Kantor, Yoav Katz, Leshem Choshen, Edo Cohen-Karlik, Naftali Liberman,
  Assaf Toledo, Amir Menczel, and Noam Slonim. 2019.
\newblock \href {https://doi.org/10.18653/v1/W19-4414} {Learning to combine
  grammatical error corrections}.
\newblock In \emph{Proceedings of the Fourteenth Workshop on Innovative Use of
  NLP for Building Educational Applications}, pages 139--148, Florence, Italy.
  Association for Computational Linguistics.

\bibitem[{Kiyono et~al.(2019)Kiyono, Suzuki, Mita, Mizumoto, and
  Inui}]{kiyono2019empirical}
Shun Kiyono, Jun Suzuki, Masato Mita, Tomoya Mizumoto, and Kentaro Inui. 2019.
\newblock \href {https://doi.org/10.18653/v1/D19-1119} {An empirical study of
  incorporating pseudo data into grammatical error correction}.
\newblock In \emph{Proceedings of the 2019 Conference on Empirical Methods in
  Natural Language Processing (EMNLP-IJCNLP)}, pages 1236--1242, Hong Kong,
  China. Association for Computational Linguistics.

\bibitem[{Li et~al.(2018{\natexlab{a}})Li, Cai, He, and
  Zhao}]{li-etal-2018-seq2seq}
Zuchao Li, Jiaxun Cai, Shexia He, and Hai Zhao. 2018{\natexlab{a}}.
\newblock \href {https://www.aclweb.org/anthology/C18-1271} {Seq2seq dependency
  parsing}.
\newblock In \emph{Proceedings of the 27th International Conference on
  Computational Linguistics}, pages 3203--3214, Santa Fe, New Mexico, USA.
  Association for Computational Linguistics.

\bibitem[{Li et~al.(2018{\natexlab{b}})Li, He, Cai, Zhang, Zhao, Liu, Li, and
  Si}]{li-etal-2018-unified}
Zuchao Li, Shexia He, Jiaxun Cai, Zhuosheng Zhang, Hai Zhao, Gongshen Liu,
  Linlin Li, and Luo Si. 2018{\natexlab{b}}.
\newblock \href {https://doi.org/10.18653/v1/D18-1262} {A unified syntax-aware
  framework for semantic role labeling}.
\newblock In \emph{Proceedings of the 2018 Conference on Empirical Methods in
  Natural Language Processing}, pages 2401--2411, Brussels, Belgium.
  Association for Computational Linguistics.

\bibitem[{Li et~al.(2020)Li, Wang, Chen, Utiyama, Sumita, Zhang, and
  Zhao}]{DBLP:conf/iclr/LiWCUS0Z20}
Zuchao Li, Rui Wang, Kehai Chen, Masao Utiyama, Eiichiro Sumita, Zhuosheng
  Zhang, and Hai Zhao. 2020.
\newblock \href {https://openreview.net/forum?id=S1efxTVYDr} {Data-dependent
  gaussian prior objective for language generation}.
\newblock In \emph{8th International Conference on Learning Representations,
  {ICLR} 2020, Addis Ababa, Ethiopia, April 26-30, 2020}. OpenReview.net.

\bibitem[{Li et~al.(2021)Li, Zhang, Zhao, Wang, Chen, Utiyama, and
  Sumita}]{DBLP:journals/corr/abs-2102-05951}
Zuchao Li, Zhuosheng Zhang, Hai Zhao, Rui Wang, Kehai Chen, Masao Utiyama, and
  Eiichiro Sumita. 2021.
\newblock \href {http://arxiv.org/abs/2102.05951} {Text compression-aided
  transformer encoding}.
\newblock \emph{CoRR}, abs/2102.05951.

\bibitem[{Lichtarge et~al.(2019)Lichtarge, Alberti, Kumar, Shazeer, Parmar, and
  Tong}]{lichtarge-etal-2019-corpora}
Jared Lichtarge, Chris Alberti, Shankar Kumar, Noam Shazeer, Niki Parmar, and
  Simon Tong. 2019.
\newblock \href {https://doi.org/10.18653/v1/N19-1333} {Corpora generation for
  grammatical error correction}.
\newblock In \emph{Proceedings of the 2019 Conference of the North {A}merican
  Chapter of the Association for Computational Linguistics: Human Language
  Technologies, Volume 1 (Long and Short Papers)}, pages 3291--3301,
  Minneapolis, Minnesota. Association for Computational Linguistics.

\bibitem[{Liu et~al.(2019)Liu, Ott, Goyal, Du, Joshi, Chen, Levy, Lewis,
  Zettlemoyer, and Stoyanov}]{liu2019roberta}
Yinhan Liu, Myle Ott, Naman Goyal, Jingfei Du, Mandar Joshi, Danqi Chen, Omer
  Levy, Mike Lewis, Luke Zettlemoyer, and Veselin Stoyanov. 2019.
\newblock Roberta: A robustly optimized bert pretraining approach.
\newblock \emph{arXiv preprint arXiv:1907.11692}.

\bibitem[{Ma and Hovy(2016)}]{ma-hovy-2016-end}
Xuezhe Ma and Eduard Hovy. 2016.
\newblock \href {https://doi.org/10.18653/v1/P16-1101} {End-to-end sequence
  labeling via bi-directional {LSTM}-{CNN}s-{CRF}}.
\newblock In \emph{Proceedings of the 54th Annual Meeting of the Association
  for Computational Linguistics (Volume 1: Long Papers)}, pages 1064--1074,
  Berlin, Germany. Association for Computational Linguistics.

\bibitem[{Maddison et~al.(2014)Maddison, Tarlow, and
  Minka}]{DBLP:conf/nips/MaddisonTM14}
Chris~J. Maddison, Daniel Tarlow, and Tom Minka. 2014.
\newblock \href
  {https://proceedings.neurips.cc/paper/2014/hash/309fee4e541e51de2e41f21bebb342aa-Abstract.html}
  {A* sampling}.
\newblock In \emph{Advances in Neural Information Processing Systems 27: Annual
  Conference on Neural Information Processing Systems 2014, December 8-13 2014,
  Montreal, Quebec, Canada}, pages 3086--3094.

\bibitem[{Napoles et~al.(2015)Napoles, Sakaguchi, Post, and
  Tetreault}]{napoles-etal-2015-ground}
Courtney Napoles, Keisuke Sakaguchi, Matt Post, and Joel Tetreault. 2015.
\newblock \href {https://doi.org/10.3115/v1/P15-2097} {Ground truth for
  grammatical error correction metrics}.
\newblock In \emph{Proceedings of the 53rd Annual Meeting of the Association
  for Computational Linguistics and the 7th International Joint Conference on
  Natural Language Processing (Volume 2: Short Papers)}, pages 588--593,
  Beijing, China. Association for Computational Linguistics.

\bibitem[{Napoles et~al.(2017)Napoles, Sakaguchi, and
  Tetreault}]{napoles-etal-2017-jfleg}
Courtney Napoles, Keisuke Sakaguchi, and Joel Tetreault. 2017.
\newblock \href {https://www.aclweb.org/anthology/E17-2037} {{JFLEG}: A fluency
  corpus and benchmark for grammatical error correction}.
\newblock In \emph{Proceedings of the 15th Conference of the {E}uropean Chapter
  of the Association for Computational Linguistics: Volume 2, Short Papers},
  pages 229--234, Valencia, Spain. Association for Computational Linguistics.

\bibitem[{Ng et~al.(2014)Ng, Wu, Briscoe, Hadiwinoto, Susanto, and
  Bryant}]{ng-etal-2014-conll}
Hwee~Tou Ng, Siew~Mei Wu, Ted Briscoe, Christian Hadiwinoto, Raymond~Hendy
  Susanto, and Christopher Bryant. 2014.
\newblock \href {https://doi.org/10.3115/v1/W14-1701} {The {C}o{NLL}-2014
  shared task on grammatical error correction}.
\newblock In \emph{Proceedings of the Eighteenth Conference on Computational
  Natural Language Learning: Shared Task}, pages 1--14, Baltimore, Maryland.
  Association for Computational Linguistics.

\bibitem[{Nicholls(2003)}]{nicholls2003cambridge}
Diane Nicholls. 2003.
\newblock The cambridge learner corpus: Error coding and analysis for
  lexicography and elt.
\newblock In \emph{Proceedings of the Corpus Linguistics 2003 conference},
  volume~16, pages 572--581.

\bibitem[{Omelianchuk et~al.(2020)Omelianchuk, Atrasevych, Chernodub, and
  Skurzhanskyi}]{omelianchuk-etal-2020-gector}
Kostiantyn Omelianchuk, Vitaliy Atrasevych, Artem Chernodub, and Oleksandr
  Skurzhanskyi. 2020.
\newblock \href {https://doi.org/10.18653/v1/2020.bea-1.16} {{GECT}o{R} {--}
  grammatical error correction: Tag, not rewrite}.
\newblock In \emph{Proceedings of the Fifteenth Workshop on Innovative Use of
  NLP for Building Educational Applications}, pages 163--170, Seattle, WA, USA,
  Online. Association for Computational Linguistics.

\bibitem[{Parnow et~al.(2020)Parnow, Li, and Zhao}]{DBLP:conf/ialp/ParnowLZ20}
Kevin Parnow, Zuchao Li, and Hai Zhao. 2020.
\newblock \href {https://doi.org/10.1109/IALP51396.2020.9310498} {Grammatical
  error correction: More data with more context}.
\newblock In \emph{International Conference on Asian Language Processing,
  {IALP} 2020, Kuala Lumpur, Malaysia, December 4-6, 2020}, pages 24--29.
  {IEEE}.

\bibitem[{Raheja and Alikaniotis(2020)}]{raheja-alikaniotis-2020-adversarial}
Vipul Raheja and Dimitris Alikaniotis. 2020.
\newblock \href {https://doi.org/10.18653/v1/2020.findings-emnlp.275}
  {{A}dversarial {G}rammatical {E}rror {C}orrection}.
\newblock In \emph{Findings of the Association for Computational Linguistics:
  EMNLP 2020}, pages 3075--3087, Online. Association for Computational
  Linguistics.

\bibitem[{Sutskever et~al.(2014)Sutskever, Vinyals, and
  Le}]{DBLP:conf/nips/SutskeverVL14}
Ilya Sutskever, Oriol Vinyals, and Quoc~V. Le. 2014.
\newblock \href
  {https://proceedings.neurips.cc/paper/2014/hash/a14ac55a4f27472c5d894ec1c3c743d2-Abstract.html}
  {Sequence to sequence learning with neural networks}.
\newblock In \emph{Advances in Neural Information Processing Systems 27: Annual
  Conference on Neural Information Processing Systems 2014, December 8-13 2014,
  Montreal, Quebec, Canada}, pages 3104--3112.

\bibitem[{Tajiri et~al.(2012)Tajiri, Komachi, and
  Matsumoto}]{tajiri-etal-2012-tense}
Toshikazu Tajiri, Mamoru Komachi, and Yuji Matsumoto. 2012.
\newblock \href {https://www.aclweb.org/anthology/P12-2039} {Tense and aspect
  error correction for {ESL} learners using global context}.
\newblock In \emph{Proceedings of the 50th Annual Meeting of the Association
  for Computational Linguistics (Volume 2: Short Papers)}, pages 198--202, Jeju
  Island, Korea. Association for Computational Linguistics.

\bibitem[{Vaswani et~al.(2017)Vaswani, Shazeer, Parmar, Uszkoreit, Jones,
  Gomez, Kaiser, and Polosukhin}]{vaswani2017attention}
Ashish Vaswani, Noam Shazeer, Niki Parmar, Jakob Uszkoreit, Llion Jones,
  Aidan~N Gomez, {\L}ukasz Kaiser, and Illia Polosukhin. 2017.
\newblock Attention is all you need.
\newblock In \emph{Advances in neural information processing systems}, pages
  5998--6008.

\bibitem[{Yang et~al.(2019)Yang, Dai, Yang, Carbonell, Salakhutdinov, and
  Le}]{DBLP:conf/nips/YangDYCSL19}
Zhilin Yang, Zihang Dai, Yiming Yang, Jaime~G. Carbonell, Ruslan Salakhutdinov,
  and Quoc~V. Le. 2019.
\newblock \href
  {https://proceedings.neurips.cc/paper/2019/hash/dc6a7e655d7e5840e66733e9ee67cc69-Abstract.html}
  {Xlnet: Generalized autoregressive pretraining for language understanding}.
\newblock In \emph{Advances in Neural Information Processing Systems 32: Annual
  Conference on Neural Information Processing Systems 2019, NeurIPS 2019,
  December 8-14, 2019, Vancouver, BC, Canada}, pages 5754--5764.

\bibitem[{Yannakoudakis et~al.(2011)Yannakoudakis, Briscoe, and
  Medlock}]{yannakoudakis-etal-2011-new}
Helen Yannakoudakis, Ted Briscoe, and Ben Medlock. 2011.
\newblock \href {https://www.aclweb.org/anthology/P11-1019} {A new dataset and
  method for automatically grading {ESOL} texts}.
\newblock In \emph{Proceedings of the 49th Annual Meeting of the Association
  for Computational Linguistics: Human Language Technologies}, pages 180--189,
  Portland, Oregon, USA. Association for Computational Linguistics.

\bibitem[{Zhang et~al.(2020{\natexlab{a}})Zhang, Chen, Wang, Utiyama, Sumita,
  Li, and Zhao}]{DBLP:conf/iclr/0001C0USLZ20}
Zhuosheng Zhang, Kehai Chen, Rui Wang, Masao Utiyama, Eiichiro Sumita, Zuchao
  Li, and Hai Zhao. 2020{\natexlab{a}}.
\newblock \href {https://openreview.net/forum?id=Byl8hhNYPS} {Neural machine
  translation with universal visual representation}.
\newblock In \emph{8th International Conference on Learning Representations,
  {ICLR} 2020, Addis Ababa, Ethiopia, April 26-30, 2020}. OpenReview.net.

\bibitem[{Zhang et~al.(2020{\natexlab{b}})Zhang, Wu, Zhao, Li, Zhang, Zhou, and
  Zhou}]{DBLP:conf/aaai/0001WZLZZZ20}
Zhuosheng Zhang, Yuwei Wu, Hai Zhao, Zuchao Li, Shuailiang Zhang, Xi~Zhou, and
  Xiang Zhou. 2020{\natexlab{b}}.
\newblock \href {https://aaai.org/ojs/index.php/AAAI/article/view/6510}
  {Semantics-aware {BERT} for language understanding}.
\newblock In \emph{The Thirty-Fourth {AAAI} Conference on Artificial
  Intelligence, {AAAI} 2020, New York, NY, USA, February 7-12, 2020}, pages
  9628--9635. {AAAI} Press.

\bibitem[{Zhao et~al.(2019)Zhao, Wang, Shen, Jia, and Liu}]{zhao2019improving}
Wei Zhao, Liang Wang, Kewei Shen, Ruoyu Jia, and Jingming Liu. 2019.
\newblock \href {https://doi.org/10.18653/v1/N19-1014} {Improving grammatical
  error correction via pre-training a copy-augmented architecture with
  unlabeled data}.
\newblock In \emph{Proceedings of the 2019 Conference of the North {A}merican
  Chapter of the Association for Computational Linguistics: Human Language
  Technologies, Volume 1 (Long and Short Papers)}, pages 156--165, Minneapolis,
  Minnesota. Association for Computational Linguistics.

\end{thebibliography}

\end{document}